\documentclass{article}

\usepackage{arxiv}

\usepackage[utf8]{inputenc} 
\usepackage[T1]{fontenc}    
\usepackage{hyperref}       
\usepackage{url}            
\usepackage{booktabs}       
\usepackage{amsfonts}       
\usepackage{nicefrac}       
\usepackage{microtype}      
\usepackage{graphicx}
\usepackage[square,numbers]{natbib}
\usepackage{doi}
\usepackage{authblk}
\title{Machine learning applications using diffusion tensor imaging of human brain: A PubMed literature review}

\author[1]{Ashirbani Saha}
\author[1]{Pantea Fadaiefard}
\author[1]{Jessica E. Rabski}
\author[2]{Alireza Sadeghian}
\author[1,3]{Michael D. Cusimano}

\affil[1]{Injury Prevention Research Office, St. Michael’s Hospital, Toronto, ON, Canada}
\affil[2]{Department of Computer Science, Ryerson University, Toronto, ON, Canada}
\affil[3]{Division of Neurosurgery, Department of Surgery, University of Toronto, Toronto, ON, Canada}

\date{}

\renewcommand{\shorttitle}{\textit{arXiv} Template}

\hypersetup{
pdftitle={A template for the arxiv style},
pdfsubject={q-bio.NC, q-bio.QM},
pdfauthor={Ashirbani Saha},
pdfkeywords={Diffusion Tensor Imaging, Feature Extraction, Machine Learning, Deep Learning, Support Vector Machine, Classification},
}

\begin{document}
\maketitle

\begin{abstract}
	We performed a PubMed search to find 148 papers published between January 2010 and December 2019 related to human brain, Diffusion Tensor Imaging (DTI), and Machine Learning (ML). The studies focused on healthy cohorts (n = 15), mental health disorders (n = 25), tumor (n = 19), trauma (n = 5), dementia (n = 24), developmental disorders (n = 5), movement disorders (n = 9), other neurological disorders (n = 27), miscellaneous non-neurological disorders, or without stating the disease of focus (n = 7), and multiple combinations of the aforementioned categories (n = 12). Classification of patients using information from DTI stands out to be the most commonly (n = 114) performed ML application. A significant number (n = 93) of studies used support vector machines (SVM) as the preferred choice of ML model for classification. A significant portion (31/44) of publications in the recent years (2018-2019) continued to use SVM, support vector regression, and random forest which are a part of traditional ML. Though many types of applications across various health conditions (including healthy) were conducted, majority of the studies were based on small cohorts (less than 100) and did not conduct independent/external validation on test sets. 
\end{abstract}


\section{Introduction}
Machine learning (ML) applications in medicine are being increasingly developed with the advancements made in deep learning. Numerous applications are developed using medical imaging~\cite{Litjens2017,Mazurowski2019}, electronic health records~\cite{Shickel2018}, and workflow data~\cite{Wang2018a,Kim2017}, in addition to internet of things for healthcare applications~\cite{Baker2017}.  Despite multiple criticisms surrounding the application of Artificial Intelligence (AI) in medicine~\cite{Emanuel2019}, academic peer-reviewed contributions of ML applications in medicine are demonstrating significant future potential and raising hopes of the medical community~\cite{Toh2019}.

In this review article, we turn our focus to ML applications of diffusion tensor imaging (DTI) of the human brain. Since its introduction in 1994, DTI analysis had been a part of different neuroimaging studies related to several brain diseases like Alzheimer’s disease~\cite{Kantarci2017,Brueggen2017}, traumatic brain injury~\cite{Wallace2018}, and depression~\cite{Tymofiyeva2017,Ugwu2015}. Due to the sensitivity of DTI towards microstructural tissues, it has gained popularity for the evaluation of White Matter (WM) architecture in both healthy and diseased adult and pediatric populations \cite{Ranzenberger2018,Douglas2015,Blamire2018,Rathore2017}. DTI has also been used as the part of the multimodal magnetic resonance imaging (MRI) analysis or neuroimaging analysis. Various review studies related to neuroimaging and ML have are present in the literature~\cite{Rathore2017,Davatzikos2019,Mateos-Perez2018,Pellegrini2018,Jo2019,Lotan2019}. However, the usage of DTI in the light of its diverse ML applications for diseased and healthy populations has not yet been studied. Despite the limitations of DTI~\cite{Farquharson2013}, it continues to be used in data-driven neuroimaging studies for developing ML applications. 

In this paper, we review the neuroimaging studies indexed over the past decade in PubMed that involved the development of ML applications using human brain DTIs. We highlight each study’s application areas, ML tasks performed, reference standards,  accomplishments, and any translational barriers.

\section{Machine Learning (ML)}
ML techniques are used to process input training data to learn patterns from the input data which represents experience to gain expertise through a computer program. The input data can be of various types e.g., imaging, text, continuous or discrete variables or a combination of these types. The training is performed using mathematical operations on the input data, such that trained model can be used to perform a specific task in the form of prediction. The mathematical operations for training are a part of the model or algorithm which is a well-defined process that enables learning the parameters/weights using the input data. Various applications of ML exist in the field of imaging analysis, text analysis, bioinformatics, and finance to name a few. Some applications of ML using medical imaging as input  are detection of intracranial haemorrhage in CT images~\cite{Chilamkurthy2018}, classification of migraine patients into different categories  using MRI~\cite{Yang2018}, cell detection in microscopic imaging~\cite{Xie2015a}. Examples of successful and popular ML algorithms include support vector machines (SVM), random forest (RF), artificial neural networks (ANN), Gaussian Mixture Model (GMM), Elastic Net Regression.  Readers interested in ML terminology terms are referred to study by Liu et al.~\cite{Liu2019} for further details on ML applications within medicine.

With the advent of Deep Learning (DL) techniques, ML applications using medical images have gathered momentum in the recent years. Neuroimaging has been a great topic of interest for the ML community over several years. Existing reviews of ML related neuroimaging include a general review of progress and challenges in neuroimaging related ML~\cite{Davatzikos2019}, use of structural neuroimaging as a clinical predictor~\cite{Mateos-Perez2018}, neuroimaging based classification and related feature extraction~\cite{Rathore2017},  and systematic reviews of ML and neuroimaging studies for diseases like Alzheimer’s~\cite{Pellegrini2018,Jo2019}. Other review articles of ML applications involving specific diseases are related to glioma~\cite{Lotan2019}, pituitary tumors~\cite{Saha2020}, Parkinson’s disease~\cite{Belic2019}, neurodegenerative diseases~\cite{Noor2019}, and schizophrenia~\cite{DeFilippis2019}. A study focusing on ML-based tractography processed using diffusion imaging was conducted by Poulin et al.~\cite{Poulin2019}.
 
Owing to a long presence of DTI for neuroimaging research, we considered applications of ML that use DTI of human brain with or without other forms of imaging and non-imaging data. We reviewed PubMed literature from the last 10 years and focused on the spectrum of types of ML tasks that leveraged usage of DTI for various health conditions.

\section{Diffusion Tensor Imaging: Basics and Usage}

DTI is a form of diffusion weighted imaging (DWI)~\cite{Alexander2007,Soares2013}. DWI is a variant of magnetic resonance imaging (MRI) that produces images by using the diffusion rate of water within brain tissue and provides elaborate details about tissue microstructure. DWI can be acquired through existing MRI scanners thereby allowing for non-invasive acquisition; and does not require the use of contrast agents or radiotracers. In plain DWIs, diffusion is modeled by a scalar parameter or diffusion coefficient. However, as diffusion is three-dimensional phenomena, it can be modeled through diffusion tensor techniques (like DTI) which accounts for difference in diffusion of water and tissue molecules by providing information about the structural orientation and quantitative anisotropy of the molecules. Details of data acquisition and processing are elaborated by Le Bihan et al.~\cite{LeBihan2001} and Soares et al.~\cite{Soares2013}. The pipeline from the acquisition to the usage of DTI is complex and the typical steps are described below:
\begin{itemize}
\item[A.] Acquisition: For DTI modeling, DWI sequences are captured through axial (typical) slices (allowing zero gaps between the slices) of the entire brain at a low gradient pulse (usually b = 0 s/sq. mm) and at least in six non-collinear diffusion encoding directions (usually b = 800-1000 s/sq. mm). As in any MRI acquisition, the acquired signal is affected by several other acquisition parameters including, but limited to, field strength, field-of-view (FOV), slice thickness, echo time, and repetition time.
\item[B.] Pre-processing and quality checks: Two major components involve: (i) visual verification of DWIs to note artifacts using general purpose image viewers, assuring the correctness of the acquisition parameters after importing the images, and conversion of raw image to specific image formats, and (ii) correction of eddy current and motion, performed as DWI sequences are characterized by low signal-to-noise (SNR) and motion artifacts. Additional and optional preprocessing can involve the removal of the skull, and other non-brain areas such as, muscle, skin, cerebrospinal fluid (CSF). etc. from the brain, registration, and normalization of the sequences.
\item[C.] Processing: The heart of DTI processing lies in the estimation of diffusion tensor, characterized by the eigenvalues and eigenvectors, at each voxel. This is often followed by the estimation of various diffusion indices, metrics, or parameters such as fractional anisotropy (FA), mean diffusivity (MD), radial diffusivity (RD), axial diffusivity (AD). DTI also enables the three-dimensional estimation (using the principal eigenvector) of the trajectory and location of white matter tracts through a mechanism called tractography. Tractography can be used to create brain-wide mapping of neuronal connections among the anatomical regions (obtained automatically/semi-automatically/manually using structural MRI) of the brain to obtain a connectome.
\item[D.] Quantitative Analysis: These analyses are often guided by the processing performed in step C. Summary measures of selected DTI indices can be extracted using anatomical regions or tracts for detecting groupwise voxel analysis such as tract-based spatial statistics (TBSS). Analysis based on voxel-by-voxel analysis (such as voxel-based analysis or VBA) are also performed. Another set of analysis includes the extraction of features (DTI indices of WM or connectomes from selected regions) which subsequently undergo the application of ML methods.
Several standard software libraries and tools are available to perform the post-acquisition stages related to DTI processing~\cite{Haddad2019,Hasan2011,Liu2015}. 
\end{itemize}
\section{Materials and Methods}

\subsection{Search Strategy and Inclusion Criteria}
We searched PubMed for ML related applications of DTI published from January 1, 2010 to December 31, 2019. The time-range was selected to review the prior decade of ML related publications using DTI. Three search queries (detailed description of the queries can be found in our supplementary material) having combination of terms “diffusion tensor imaging” and several DTI related terms and “machine learning”, “deep learning” and other ML-related terms were executed in PubMed to retrieve unique articles in the past decade. As shown in Figure 1, we included 148 articles in this review after screening and excluding 323 articles. Though, ours is a literature review, we followed the steps of PRISMA for inclusion of studies in our review. Studies included applied ML to real (in contrast to only simulated data) DTI of human brain. Our study excluded articles if usage of DTI was not confirmed and this included all the studies involving application of several other diffusion imaging acquisition techniques including high-angular diffusion imaging (HARDI). As there is inconsistency in the description of exact criteria for classifying a diffusion MRI acquisition as a HARDI acquisition~\cite{Mueller2015} as opposed to other diffusion MRI acquisitions such as diffusion kurtosis imaging, diffusion spectrum imaging; we excluded works involving only HARDI acquisition and focused on DTI acquisition.
From the studies reviewed, we noted the (a) first author and year of publication, (b) unmet need addressed by the study, (c) datasets and ML tasks performed, (d) reference standard, (d) usage of DTI in the study, (e) ML model used and; (f) performance of the models with accomplishments and limitations of the studies. We categorized the studies based on the cohorts that they use for the design of their ML problems. Figure 1 shows a graph of the number of studies published per year. 
\begin{figure}
  \includegraphics[width=0.9\textwidth]{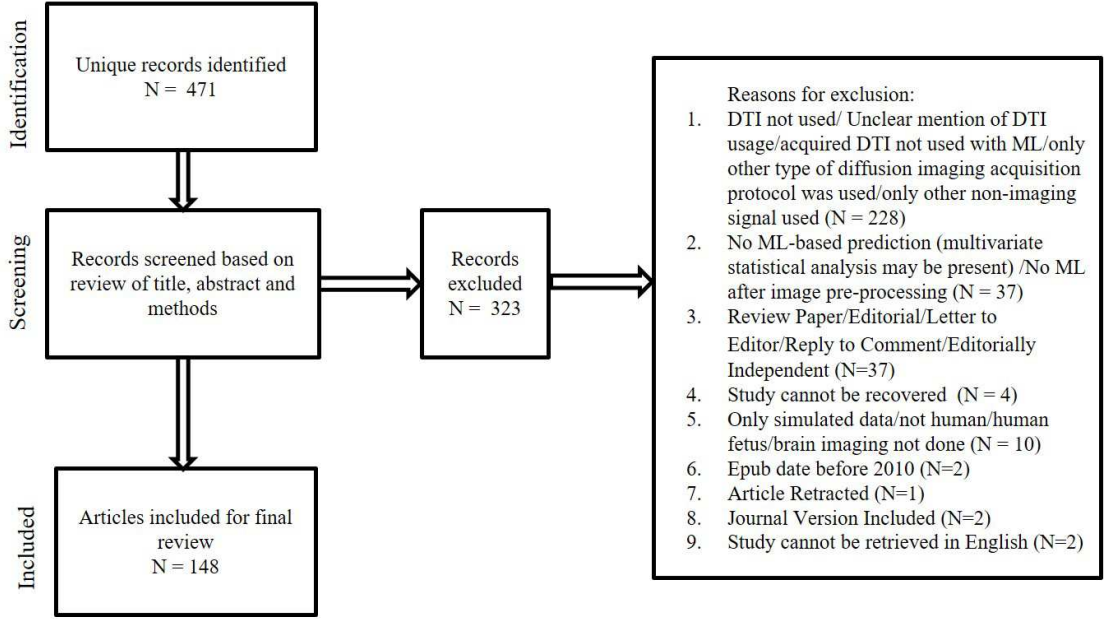}
  \caption{Consort diagram for the studies reviewed}
  \label{fig:consort_diagram}
\end{figure}
The following categories were found from the studies included in our work: 
\begin{itemize}
\item[A.] Studies (n = 15) with fully healthy cohorts 
\item[B.] Studies (n = 121) that included individual cohorts that may have an addition of healthy controls/relevant control along with the following types of disorders or conditions 
\begin{itemize}
\item[1.] mental health disorders (n = 25)
\item[2.] tumor (n = 19)
\item[3.] trauma (n = 5)
\item[4.] dementia (n =24)
\item[5.] developmental disorders (n = 5)
\item[6.] movement disorders (n = 9)
\item[7.] other neurological disorders (n = 27)
\item[8.] miscellaneous non-neurological disorders or did not state the disease of focus (n = 7)
\end{itemize}
\item[C.] Studies (n = 12) that had multiple cohorts with characteristics of A or B or mixed cohorts involving more than one disease type in B
\end{itemize}
We discuss the specifics of these studies in the next section. Further details of these studies can be found in our supplementary material (sTables 1- 10)

\section{Results}
\subsection{Studies with Fully Healthy Cohorts} 
Of the 15 articles that involved healthy subjects, three focused on parcellation of specific type of brain areas such as cortex~\cite{Gorbach2018}, left-dorsal pre-motor cortex~\cite{Genon2018} and supplementary motor cortex and pre-supplementary motor cortex~\cite{Crippa2011}. Tissue segmentation was done in two studies~\cite{Wen2013,Ciritsis2018} where the first one focused on the robustness of the technique for noise and field inhomogeneities. Of the four studies working on brain aging, one study~\cite{Robinson2010} classified the patients into age groups and the other three works~\cite{Kawahara2017,Lin2016,Mwangi2013} studied the regression of healthy human brain with age. Additionally, the regression of cognitive performance scores was performed in two studies to predict cognitive ability~\cite{Yun2013} and memory assessments~\cite{Ullman2014}. Other applications include: denoising~\cite{Maximov2012}, network analysis of brain connectivity~\cite{Wang2014}, white matter patterns in infants~\cite{Hua2012}, and zero-shot learning of fMRI task using DTIs~\cite{Fuchigami2018}. The usage of DTI for all the applications mentioned in this section lied in extracting DTI-parameters, performing DTI based tractography, and estimation of connectivity matrices. Only six of these studies~\cite{Genon2018,Kawahara2017,Lin2016,Mwangi2013,Ullman2014,Hua2012} used a dataset involving more than 100 participants and highest number of subjects was 323 used in the study by Ullman et al.~\cite{Ullman2014}.

The ML tasks of the studies varied, and various study-specific reference standards were collected. The ML models designed/used for the applications varied accordingly. Nine studies focused on the novelty of the overall ML approach used such as innovations on convolutional filters used in artificial neural networks (ANN)~\cite{Kawahara2017}, combination of ANN with hybrid genetic algorithm~\cite{Lin2016}, novel modeling of noise distribution~\cite{Maximov2012}, multigraph min-max cut~\cite{Wang2014}, and modified Bayesian modeling~\cite{Ullman2014}, and incorporation of DTI-based registration in a zero-shot support vector machine (SVM)-based prediction of fMRI task~\cite{Fuchigami2018}. Existing supervised ML techniques such as SVM, maximum uncertainty linear discriminant analysis (LDA), relevance vector machine, linear regression, partial least squares, Multilayer perceptron (MLP) were used by four studies~\cite{Ciritsis2018,Robinson2010,Mwangi2013,Yun2013}. Unsupervised ML approaches such as K-means clustering, force directed graph layout, fuzzy C-means clustering with spatial constraints were used by three studies~\cite{Genon2018,Crippa2011,Wen2013} respectively. The study by Gorbach et al.~\cite{Gorbach2018} validated pipelines for connectivity-based cortical parcellation. Critsis et al.~\cite{Ciritsis2018} evaluated the performance of their model in an validation set with an accuracy of 82.1\%. Wen et al.~\cite{Wen2013} validated the clustering methodology for segmentation with a volume overlap over 0.84 WM, gray matter (GM), and CSF.

\subsection{Mental Health Disorders}

We included 25 studies which focused on various mental health disorders: Autism Spectrum Disorder (ASD) (N=4), first episode of schizophrenia-spectrum disorder (FES) or schizophrenia (SZ) (N=5), depression disorders (N=10), First Episode Psychosis (FEP) (N=2), attention deficit hyperactivity disorder (ADHD) (N=1), obsessive compulsive disorder (OCD) (N=2), and anorexia nervosa (AN) (N=1). 
Studies on depression disorders developed classification systems for bipolar disorder (BD), major depressive disorder (MDD), and healthy controls (HC)~\cite{Deng2018a}, classification of BD patients and controls to data driven phenotypes~\cite{Wu2017} and classification of pediatric BD patients and controls~\cite{Mwangi2015}. Classification of MDD and controls was performed in four studies~\cite{Chu2018a,Qin2014,Fang2012,Bi2016}. Classification among acute, remitted MDD and healthy controls by Qin et al.~\cite{Qin2015}, between treated and untreated MDD by Chu et al.~\cite{Chu2018b}, and between treatment responses by in a multi-site study by Wang et al.~\cite{Wang2019} were performed. Studies on SZ focused on classification of FES and controls~\cite{Deng2019,Mikolas2018}  and of drug naïve FES and controls~\cite{Zhuang2019}, and clustering of SZ and controls~\cite{Zwir2015,Ingalhalikar2012}. The study by Ingalhalikar et al.~\cite{Ingalhalikar2012} also performed the clustering in a cohort of ASD and controls. Other studies involving ASD include classification of ASD risk in infants~\cite{Jin2015}, classification of ASD and typically developing controls~\cite{Ingalhalikar2011} and regression of ASD-related interview scores~\cite{Fouque2011}. Two studies included classification of FEP patients and controls~\cite{Peruzzo2015,Pettersson-Yeo2013}. Other studies performed classification between healthy controls and other mental health conditions like AN~\cite{Pfuhl2016}, OCD~\cite{Li2014a,Mas2016}, and ADHD~\cite{Chaim-Avancini2017}. Apart from the studies by Foque et al.~\cite{Fouque2011}, regression was performed in two studies for separate purposes: chlorpromazine dose and symptom levels in FES patients~\cite{Mikolas2018} and general psychopathology levels and age~\cite{Alnaes2018} on publicly available neurodevelopmental cohort. DTI was used to extract study-specific DTI-parameters (sometimes from WM/TBSS skeletons), perform DTI-based tractography, and estimate connectivity matrices. Connectivity matrices were often used for further feature extraction used in ML models. Eight of these studies~\cite{Deng2018a,Wu2017,Wang2019,Deng2019,Mikolas2018,Pfuhl2016,Chaim-Avancini2017,Alnaes2018} used a dataset size of more than 100, with the largest dataset (729 studies) used in the study by Alnæs et al.~\cite{Alnaes2018}.

SVMs and their variations were used for classification in 21 of the 25 mental health-related studies and Support Vector Regression (SVR) was used for regression in one of the studies~\cite{Mikolas2018}. The remaining studies focused on Random Forest (RF)~\cite{Deng2019} and least absolute shrinkage and selection operator (LASSO) and  Elastic Net~\cite{Wu2017} for classification. For clustering, variant of Non-negative matrix factorization (NMF), spectral clustering with modifications~\cite{Zwir2015,Ingalhalikar2012},  K-Means clustering~\cite{Wu2017,Pfuhl2016} were used. Of the three studies which involved regression~\cite{Mikolas2018,Fouque2011,Alnaes2018} , two studies used Kernel Ridge Regression~\cite{Mikolas2018,Fouque2011}  and one study used shrinkage linear regression~\cite{Alnaes2018}. One study by Deng et al.\cite{Deng2019} validated their model in a hold-out set with an accuracy of 76\%.

\subsection{Tumor}
Of the 19 studies involving tumors, 18 were related to brain tumors, and one was related to nasopharyngeal carcinomas~\cite{Leng2019} which had a prospective design for the classification of different stages of post-therapy time after the patients were treated for radiotherapy. Fourteen of the studies focused on the following aspects of glioma brain tumors specifically:  multimodal segmentation of glioma~\cite{Soltaninejad2018}, classification related to tumor biology (i.e. tumor grade)~\cite{Shibata2014,Vamvakas2019,Citak-Er2018}, IDH1 or/and MGMT status prediction~\cite{Eichinger2017,Chen2017a,Chen2018}, nuclei-content classification~\cite{Hu2015}, EGFRvIII status prediction~\cite{Akbari2018}, tumor cell density regression~\cite{Hu2019}, and tissue type classification for causing recurrence~\cite{Akbari2016}, and prognostication by classifying true versus pseudo-progression~\cite{Qian2016,Zhang2016}, classification of survival period as good or bad~\cite{Liu2016}. Segmentation of different types of brain tumors and classification of high grade glioma versus solitary metastasis was investigated in the study by Yang et al.~\cite{Yang2014}. Classification of different types of intracranial lesions was performed by Svolos et al.~\cite{Tsougos2013}. Differentiation of atypical meningiomas from glioblastoma and metastases was performed by Svolos et al.~\cite{Svolos2013}. Shrot et al.~\cite{Shrot2019} performed a study for classification of four types of brain tumors. Six of the 19 studies~\cite{Leng2019,Shibata2014,Akbari2018,Zhang2016,Tsougos2013,Svolos2013} had dataset sizes of more than 100, with the study by Leng et al.~\cite{Leng2019} having the largest sample size of 147.

The usage of DTI included derivation of DTI-parameters for feature representation~\cite{Leng2019,Shibata2014,Vamvakas2019,Hu2015,Akbari2018,Hu2019,Akbari2016,Qian2016,Zhang2016,Tsougos2013,Svolos2013,Shrot2019}, extraction of features using structural connectome~\cite{Leng2019,Chen2017a,Chen2018,Liu2016} and deriving isotropic and anisotropic components for segmentation~\cite{Soltaninejad2018,Hu2015,Yang2014}. Since several studies require tumor and/or its surroundings as the regions-of-interest, (ROI), manual ROIs were used in some applications for identifying tumor or specific regions of tumor and its surroundings~\cite{Shibata2014,Hu2015,Akbari2016,Tsougos2013,Svolos2013} and manual seed-points were used in~\cite{Yang2014}. Atlas-based ROIs were used by~\cite{Leng2019,Chen2018,Liu2016}.  All studies performed classifications and SVMs were used in 13 of 17 studies that performed classification tasks. RF, quadratic discriminant analysis (QDA), LDA, naïve Bayes (NB), and artificial neural networks (ANNs) were used in some of the studies. Variants of matrix completion techniques were used as clustering approaches in one study~\cite{Chen2018} performing unsupervised prediction of IDH1 and MGMT status. Akbari et al.~\cite{Akbari2018} validated their radiomic signature for glioma in a replication cohort with an AUC of 0.86. Eichinger et al.~\cite{Eichinger2017} reported an AUC of  0.91 in validation cohort for IDH classification. Akbari et al.~\cite{Akbari2016} reported an AUC of 0.84 in replication cohort for predicting early recurrence and Hu et al.~\cite{Hu2015} reported an accuracy of 81.8\% in the validation set.

\subsection{Trauma}
Five studies involved the classification of traumatic brain patients (TBI) injury versus controls. Three of the studies~\cite{Vergara2016,Zheng2017,Sharp2012} involved patients with the presence of mTBI (mild TBI) and used varying reference standards. The remaining two studies involved patients with history of mTBI whose MRI was acquired on average around 38 months after injury~\cite{Fagerholm2015} or after retirement from sports~\cite{Goswami2016}.  DTI was used for DTI parameter extraction from white matter, brain voxels, performing tractography and for generating connectivity matrices. SVM was used in all studies. Along with classification, two studies  performed regression:  for neuro-physical outcome scores using SVR~\cite{Sharp2012} and regression for cognitive impairment scores using Elastic Net~\cite{Fagerholm2015}. Other than one study~\cite{Vergara2016} which had 100 subjects, other studies had a dataset sizes less than 100. The study by Fagerholm et al.~\cite{Fagerholm2015} reported a correlation of 0.44 in the validation set.

\subsection{Dementia}
Of the 24 dementia-related studies, one study~\cite{Feis2018} classified presymptomatic FTD mutation carriers versus controls whereas the majority (N = 12) of studies included Alzheimer’s disease (AD) patients in their cohort. The reference standards for these studies included clinically probable AD based on NINCDS-ADRDA criteria and/or clinical diagnosis of AD~\cite{Wada2019,Maggipinto2017,Bron2017,Li2014b,Dyrba2013,Chen2017b}. Some studies~\cite{Schouten2017,Dyrba2015} included additional criteria such as CSF AD markers and one study used MMSE scores and the ADNI protocol~\cite{Patil2013}. While classification of AD versus healthy controls was the focus of several studies, Bron et al.~\cite{Bron2017} developed classification systems for discriminating between AD, FTD, and controls. Maggipinto et al.~\cite{Maggipinto2017} and Eldeeb et al.~\cite{Eldeeb2018} studied application of DTI measurements for classification between healthy controls, mild cognitive impairment (MCI), and AD. MCI versus healthy classification was performed by several other studies~\cite{Ciulli2016,Diciotti2015,Pineda-Pardo2014,Zhu2014,Xie2015b,Wee2012,ODwyer2012,Lee2013} with two studies involving vascular MCI~\cite{Ciulli2016,Diciotti2015} and one study involving amnestic MCI~\cite{Knight2019}. Classification of various subtypes of MCI was studied by Haller et al.~\cite{Haller2013}. A multicenter study on the classification of two of types of MCI and healthy controls was performed by Dyrba et al.~\cite{Dyrba2015_2}. A study by O’Dwyer~\cite{ODwyer2012_2} classified apolipoprotein-E4 (a risk-factor for AD) carriers versus non-carriers in healthy young patients. In these studies DTI was used to extract the parametric values, features from white matter, white matter tracts, gray matter tracts and identified ROI, and generate connectivity matrices. Seven of these studies~\cite{Feis2018,Maggipinto2017,Dyrba2013,Schouten2017,Xie2015b,Knight2019,Bouts2019} had a dataset size of more than 100 with the highest dataset size of 915 subjects (combination of several cohorts) used in the study by Bouts et al.~\cite{Bouts2019}.

All but one of the 24 studies performed classification and SVM and its variations (binary, multi-class) were used in 18 studies for classification. Among these 18 studies, SVM as well as other classifiers including LDA, KNN, Functional Trees, NB were used by three studies~\cite{Dyrba2013,Diciotti2015,Pineda-Pardo2014}. CNN, Adaboost, RF, logistic regression (LR), elastic net regression classifier, sparse group LASSO were used by the other six studies. Two studies- Maggipinto et al.~\cite{Maggipinto2017} and Lee et al.~\cite{Lee2013} reported an AUC of 0.86 (AD vs. control classification) and accuracy of 100\% (MCI vs. control classification) respectively in corresponding validation sets.

\subsection{Developmental Disorders}
Of the five studies found regarding developmental disorders, one study focused on genetic variability in PPARG and brain connectivity~\cite{Krishnan2017} in preterm infants. Future behavioral profile in neonates was studied by Wee et al.~\cite{Wee2017}.  Classification of dyslexia versus controls in children was performed in two studies~\cite{Chimeno2014,Cui2016}.  In a set of adult patients, classification of chromosome 22q11.2 deletion syndrome versus controls was studied in the remaining study~\cite{Tylee2017}. Usage of DTI included metric extraction, tractography, and connectivity matrix generation with study-specific variations. Two studies~\cite{Krishnan2017,Wee2017} had a data size more than 100 and Krishnan et al. used a dataset of 272 pre-term infants~\cite{Krishnan2017}. SVM was used in four of the five studies. One study used Sparse reduced rank regression. Studies did not use an external validation set or were unclear about the same.

\subsection{Movement Disorders}
All the nine studies related to movement disorders involved patients with Parkinson’s disease (PD). Classification of Parkinson’s patients versus healthy controls was performed in three studies~\cite{Lei2019,Haller2012,Liu2017} with some differences in the reference standard for the disease (sTable 7 in our supplementary document). One study by Liu et al.~\cite{Liu2016Folded} involved the identification of a biomarker through feature selection using the United Kingdom brain bank criteria for diagnosis. Six studies focused~\cite{Lei2019,Jin2019,Morisi2018,Du2017,Cherubini2014a,Cherubini2014b} on several forms of Parkinsonism.  Du et al.~\cite{Du2017} reported some patients in their cohort to have post-mortem confirmed diagnosis of PD and progressive nuclear palsy (PSP). Two studies~\cite{Lei2019,Du2017} used datasets of more than 100 samples and Lei et al.~\cite{Lei2019} used a dataset of 238 subjects.

Of the eight studies that performed classification, SVMs were used in seven~\cite{Lei2019,Haller2012,Jin2019,Morisi2018,Du2017,Cherubini2014a,Cherubini2014b} and LR was used in one study~\cite{Liu2017}. One study~\cite{Liu2016Folded} performed feature selection and used Folded concave penalized model for the same. The studies reported accuracy or AUC above 0.90 for the tasks they addressed though none of these had external validation sets.

\subsection{Other Neurological Disorders}
Of the 27 studies related to other neurological disorders, majority~\cite{Munsell2019,Park2019,Mithani2019,Gleichgerrcht2018,Taylor2018,DelGaizo2017,Fang2017,Kamiya2016,Munsell2015,Pustina2015,Cantor-Rivera2015,Amarreh2014,An2014,Focke2012,Wang2018b,Paldino2014} of the studies involve epilepsy patients. Various classification tasks within temporal lobe epilepsy (TLE) patients (N=9) were performed including determination of laterality~\cite{Kamiya2016,Pustina2015,An2014}; prediction of seizure frequency~\cite{Park2019}; detection of GM and WM abnormalities of focal cortical dysplasia in extratemporal TLE~\cite{Wang2018b}; discrimination between surgical outcomes~\cite{Gleichgerrcht2018,Taylor2018,Munsell2015}, and distinguishing treatment (vagus nerve stimulation) outcomes~\cite{Mithani2019}. Among these studies, four~\cite{Park2019,Fang2017,Kamiya2016,Munsell2015} performed classification between healthy controls and TLE patients as well. A similar classification task in adult cohorts was performed by Del Gaizo et al.~\cite{DelGaizo2017} whereas children and adolescents were classified into active TLE, remitting TLE and healthy in the study by Amarreh et al.~\cite{Amarreh2014}. Classification of mTLE and controls using DTI and T2-relaxometry was performed by Cantor-Rivera et al.~\cite{Cantor-Rivera2015}.  Regression of language performance score of TLE patients was performed by Munsell et al.~\cite{Munsell2019}. Classification of language phenotypes was performed in paediatric epilepsy patients by Paldino et al.~\cite{Paldino2014} and patients of mTLE with hippocampal sclerosis (HS) were found to be different from controls based on white matter abnormalities. MR image voxel-based classification of mTLE having HS and controls was performed by Focke et al.~\cite{Focke2012}. Classification of amylotropic lateral sclerosis (ALS) versus controls using neuroimaging profile of the corticospinal tract was performed by Sarica et al.~\cite{Sarica2017}. In another study~\cite{Ferraro2017}, the usage of multimodal MRI was explored for detection and comparison of ALS, predominantly upper motor neuron disease (PUMN), ALS-mimicking conditions and controls. In two separate tasks, classification of controls and relapsing-remitting multiple sclerosis (MS) patients and classification of MS with different degree of disability was performed by Zurita et al.~\cite{Zurita2018}. Wen et al.~\cite{Wen2017} performed a study on classification of Tourette Syndrome (TS) patients and controls.  Imaging and questionnaire data were both used in the classification of migraine versus control patients in~\cite{Garcia-Chimeno2017}. Thalamus segmentation for cerebellar ataxia patients was performed by Stough et al.~\cite{Stough2014} using multimodal MRI. Clustering of HIV-infected and controls for cognitive impairment~\cite{Underwood2017}, classification of presence and absence hepatic minimal encephalopathy~\cite{Chen2015}, classification of normal versus abnormal outcomes in neo-natal encephalopathy~\cite{Ziv2013}, classification of stroke outcomes using acute DTI~\cite{Moulton2019}, and detection of sensory processing disorder using DTI tractography~\cite{Payabvash2019} are other applications. Four of the studies~\cite{Munsell2015,Ferraro2017,Zurita2018,Underwood2017} had dataset size more than 100 with the study by Ferraro et al. having the largest dataset size of 265 patients~\cite{Ferraro2017}. 

SVMs were used by 17 of the 23 studies for classification. Other ML models considered were RFs and their variants, multiple kernel learning, Adaboost, naïve Bayes, and gaussian process for ML. Two studies reported results on external validation cohorts. Mithani et al.~\cite{Mithani2019} validated their results on multi-institutional independent dataset with an accuracy of accuracy of 83.3 for predicting treatment outcome. Ferraro et al.~\cite{Ferraro2017} validated their model with an accuracy of 95\% for classifying PUMN vs. ALS mimicking disorders.

\subsection{Miscellaneous Non-neurological Disorders/Not-stated}
The diseases/conditions assigned to the studies (n = 7) in this section were related to the classification of venous erectile dysfunction patients and healthy controls~\cite{Li2018}, exploring amygdala modulation in adolescent representatives of families facing more economic and social challenges~\cite{Goetschius2019}, classification of world-class gymnasts and healthy controls using neuroplasticity in white matter~\cite{Deng2018b}, classification of self-reported drinking histories of adolescents~\cite{Park2018}. Three studies did not reveal the state (disease conditions/healthy) of the participants considered and developed methodologies for the segmentation of tissues~\cite{Wen2014} and WM tracts~\cite{Hao2013,Mu2019}. Three studies~\cite{Goetschius2019,Park2018,Mu2019} used a dataset size of more than 100;  and largest dataset (7000 samples had DTI) was used by Mu et al.~\cite{Mu2019}.

SVMs were used for classification in two studies~\cite{Li2018,Deng2018b}. Sparse logistic classification model and LR were also used for classification in two studies~\cite{Deng2018b,Park2018}. For segmentation and regression, fully convolutional networks (FCN), Bayesian decision theory, and combination of maximum a-posteriori, Markov random field, gradient descent, and kernel regression were used. These studies did not use an external validation set or were unclear about the same.

\subsection{Multiple or Mixed Cohorts}

We found 12 studies belonging to this category. FA and MD maps were reconstructed using ANNs by Aliotta et al.~\cite{Aliotta2019} using a training set of healthy patients and the trained model was tested in a cohort of GBM patients. Classification of AD and Late Onset BD was performed by Besga et al.~\cite{Besga2016} and classification of AD, late onset BD and healthy controls were performed in another study~\cite{Besga2012}. Classification of AD, non-AD in the context of corticobasal syndrome, and healthy controls were performed by Megdalia et al.~\cite{Medaglia2017}. A regression study of age in healthy humans and prediction of age gap in HIV+ and HIV- patients were performed by Kunh et al.~\cite{Kuhn2018}.  Pathway reconstruction using neuroimaging in a public cohort of healthy and schizophrenic patients was performed in the study~\cite{Chu2015}. Another study~\cite{Yendiki2016} on reconstruction of white matter pathways using longitudinal diffusion MRI was tested separately in healthy control and patients of Huntington disease for different types of analysis. Cortical ROIs were segmented in healthy and MCI patients by Zhang et al.~\cite{Zhang2013}. Clustering of brain sub-networks in healthy and ASD patients was performed~\cite{Ball2017}. Clustering using DTI connectivity matrices and  analyzing children with ASD and TDs and gender differences in different cohorts was performed by Ghanbari et al.~\cite{Ghanbari2013}. Classification of ASD and typically developing controls and regression of responsiveness scale scores of meditation practitioners was performed by Adluru et al.~\cite{Adluru2013}. Lin et al.~\cite{Lin2014} created a classification of MCI and severe dizziness in presymptomatic carotid stenosis. Four studies~\cite{Kuhn2018,Ball2017,Ghanbari2013,Adluru2013} used datasets of size more than 100 and Kuhn et al.~\cite{Kuhn2018} used a dataset size of 869 including their independent test set.

SVMs were used for the classification of five studies~\cite{Besga2016,Besga2012,Medaglia2017,Adluru2013,Lin2014} and Support vector regression was used by two studies~\cite{Kuhn2018,Adluru2013}. Clustering was performed using NMF and automatic relevance determination based projective NMF in two studies~\cite{Ball2017,Ghanbari2013}. Generalized Multiple Kernel Learning was used for ROI localization by Zhang et al.~\cite{Zhang2013} and a new technique for fiber reconstruction involving agglomerative hierarchical clustering was used for pathway reconstruction~\cite{Chu2015}. In the testing set, Kuhn et al.~\cite{Kuhn2018} predicted brain age gap which was significantly correlated (r = 0.26) with the cognitive function. Yendiki et al.~\cite{Yendiki2016} executed test-retest analysis (average error was 5\%) on a separate cohort of 9 healthy patients and sensitivity analysis (22.3\% of positions in 18 WM pathways showed significant changes) on 46 Huntington disease patients. Chu et al.~\cite{Chu2015} demonstrated a better average results (spatial metric 2.72 mm) compared to an existing technique. Zhang et al.~\cite{Zhang2013} validated their method and found its performance (prediction error: 8.08 mm) to be better than four existing techniques.

\section{Discussion}

Our study found articles relating to DTI applications for several medical disorders and conditions (including healthy). The majority of applications involved the classification of patients of various medical conditions. Many studies also included the clustering of patients, regression of clinical test scores using imaging, segmentation of tissues (normal and abnormal), and novel methods of DTI reconstruction. Classification was performed in 114 studies and SVM was used in the majority (N = 93) of the studies as the preferred classifier. A significant portion (31/44) of publications in the recent years (2018-2019) continued to use traditional ML methods such as SVM, support vector regression and random forest despite phenomenal development in deep learning. 

While performing an ML task, DTI was sometimes used in conjunction with other imaging and non-imaging data. Application of human brain DTI-based ML was not confined to subjects for having neurological disease; non-neurological disorders and healthy patients were also considered for designing several studies. Classification of various diseases and controls, clustering of diseased (with various levels of the disease) and healthy cohorts, regression of clinical test scores were major application in studies involving cohorts having abnormal health conditions with classification as the predominant topic of interest. For healthy cohorts, classification, regression, and image segmentation received balanced interest from the researchers.

The majority of the studies had datasets of less than 100 samples. Though most of the studies conducted internal validation through cross-validation, few studies conducted validation on an independent test set. Studies often stated small sample size, study design (such as retrospective, cohort studies), inconsistencies or variabilities in reference standards (often unique to the problem studied), side effects of medications, limited number of directions of diffusion, errors introduced during pre-processing of DTI, crossing-fibers, restriction of extracting imaging features in particular regions, and imbalances in the distribution of samples, as study limitations. Since validating on an independent test set is one of the tests of generalizability of the models, this needs to be addressed by majority of the studies for clinical translation.

While the studies reviewed show promise, before ML techniques can really be translated to the daily healthcare setting to  improve the diagnosis, treatment, and prognostication of patients with the neurological conditions reviewed, future work must include larger datasets to validate their accuracy and efficacy. Future work also has to more carefully consider reference standards for diagnosis as these vary within conditions, and the training and evaluation of the ML methods rely heavily on these. Moreover, since several technically varying ML solutions can be developed to solve the same task, researchers and organizations will need to collaborate to develop large scale, diverse, task-specific benchmark datasets to ensure that robust and well-performing sets of ML solutions can be developed to advance healthcare solutions. Ideally, these datasets should be made open-access to allow critical evaluation of these techniques by multiple research teams so that more rapid translation of these techniques to patient care can be achieved with confidence.

\section{Limitations}

Our study selected papers indexed in PubMed only and as such, does not include articles indexed in other research databases. However, since PubMed is representative of biomedical literature across the world, our study captures a significant view of the research involving ML application related to DTI globally. It is likely that we have excluded some articles from 2019 that were not indexed in PubMed when we executed our search. Due to non-inclusion of HARDI, some connectome related ML applications are excluded from this review. Nevertheless, various ML applications using connectome were considered in our work.

\section{Acknowledgements}
Funding support from the Canadian Institute for Military and Veteran Research (CIMVHR) is gratefully acknowledged.






\bibliography{refs}

\end{document}